\documentclass{article}

     \usepackage[final]{neurips_2019}

\usepackage[utf8]{inputenc} 
\usepackage[T1]{fontenc}    
\usepackage{hyperref}       
\usepackage{url}            
\usepackage{booktabs}       
\usepackage{amsfonts}       
\usepackage{nicefrac}       
\usepackage{microtype}      
\usepackage{subfigure}
\usepackage{moreverb}
\usepackage{epsfig}
\usepackage{amsmath,amssymb,amsthm,mathrsfs,amsfonts,dsfont}
\usepackage{adjustbox,lipsum}
\usepackage{algorithm,algorithmic}
\usepackage{amsfonts}
\usepackage{epsfig}
\usepackage{amssymb}
\usepackage{amsmath}
\usepackage{xcolor}

\title{Representation of Federated Learning via Worst-Case Robust Optimization Theory}

\author{%
	Saeedeh Parsaeefard   \,\,\,\, \,\,\,\   Iman Tabrizian    \,\,\,\ \,\,\,\ Alberto Leon Garcia
\\	Electrical and Computer Engineering\\
	University of Toronto\\
	Toronto, ON., M4P 1A6 \\
	\texttt{saeideh.fard, iman.tabrizian,  alberto.leongarcia @utoronto.ca} \\
}

\begin{document}

\maketitle

\begin{abstract}
Federated learning (FL) is a distributed learning approach where a set of end-user devices participate in the learning process by acting on their isolated local data sets. Here, we process local data sets of users where worst-case optimization theory is used to reformulate the FL problem where the impact of local data sets in training phase is considered as an uncertain function bounded in a closed uncertainty region. This representation allows us to compare the performance of FL with its centralized counterpart, and to replace the uncertain function with a concept of protection functions leading to more tractable formulation. The latter supports applying a regularization factor in each user cost function in FL to reach a better performance. We evaluated our model using the MNIST data set versus the protection function parameters, e.g., regularization factors.

\end{abstract}

\section{Introduction and Motivation}
FL is a promising approach to solve two main implementation challenges \citep{FederatedMachineLearning}: 1) increasing the amount of isolated, collected data from smart end-user devices such as IoT sensors and smart phones;
 2) privacy concerns of end-users and their reluctance to share their data with cloud owners.   
FL leverages the recent high computation and communication capabilities of these devices and aims to provide a unified framework for distributed learning among them to train a model and solve related optimization problems.

A main approach to solve FL problems is to apply the iterative algorithms \citep{distributedbook}, such as the gradient decent (GD) and its subsidiaries, e.g, stochastic GD, Newton methods \citep{gfed}, block coordinated descent (BCD) \citep{120891009,7776948}, to solve the problems via updating the weights or gradients of the related problems among users in a primal dual distributed manner. This type of algorithms usually deal with a convergence rate and the performance gap between their solutions and their counterpart centralized solutions \citep{distributedbook}. Consequently, there is always an interest to propose an algorithm with faster convergence rate and better performance gain. Recently, there has been a surge in works dealing with the aforementioned problem where perturbed or regularization terms are added to the objective functions \citep{Fedprox,Fedfair,gfed,McMahan2016CommunicationEfficientLO}. 

In this paper, we aim to study the gap between FL and its centralized counterpart solutions in order to give more insight about its behavior with regularization factors. We initiate this study by highlighting that, each user local data set in FL is a truncated or uncertain portion of the data set of a related counterpart centralized solution. Therefore, end-users' solutions are uncertain in the sense that each user does not have any knowledge about interactions between data sets of other users and its own. To model this uncertainty, we deploy worst-case robust optimization theory where this uncertain/unknown effect between data sets of users is considered as an unknown function in their objective functions. However, there is no clues about a form and mathematical representation of this unknown function. We will discuss how worst-case robust optimization theory allows us to assume a very general form for this function such as linear function; and then, model the unknown part of this function as a bounded error trapped in uncertainty region \citep{protectionfunction,saeedehbook}. Still the solution of the worst-case presentation for FD is not trivial. To handle this, we resort to the variational inequality (VI) \citep{PangVI}, and show how the solution of the proposed problem can be related to its centralized counterpart solution. Accordingly, we can study the performance gap between FL and its counterpart centralized solutions. Then, we can show how with some practical assumptions, the uncertain function for each user can be replaced by the bound of uncertainty region and weights of other users via the concept of protection functions \citep{protectionfunction,saeedehbook}. The interesting point is that this protection function has a similar structure to the perturbed or regularized factors of previous works, e.g., \citep{Fedprox}. Hence, representation of FL via worst-case robust optimization theory can support their results. 
We evaluate the performance of our model on the EMNIST data set for digit values of users \citep{cohen2017emnist} based on the bound of uncertainty region and regularization factors. Our evaluations demonstrate that linear norm function order 2 (L2) can outperform other regularization factors on this data set in a federated setting in terms of cost function.




\section{General Federated Learning Setup}
Consider a set of data owners/users/nodes/entities denoted by $\mathcal{N}=\{1,\cdots,N\}$, all of whom wish to train a machine learning model by considering their data sets  $\{\mathcal{D}_1=\{\textbf{x}_1,\textbf{y}_1\},\cdots,\mathcal{D}_N=\{\textbf{x}_N,\textbf{y}_N\}\}$. In this setup, feature $\textbf{x}_n$, label $\textbf{y}_n$, and  weights $\textbf{w}_n$ belong to user $n$ which builds the complete training set \citep{7113892}. Consider $\textbf{w}=\{\textbf{w}_1,\cdots \textbf{w}_N\}$, $\textbf{w}=\{\textbf{w}_1,\cdots \textbf{w}_N\}$, $\textbf{x}=\{\textbf{x}_1,\cdots \textbf{x}_N\}$, and $\textbf{y}=\{\textbf{y}_1,\cdots \textbf{y}_N\}$; where in this setup, the goal is to solve 
\begin{equation} \label{1}
\min_{\textbf{w} \in \textbf{W}} V(\textbf{w}, \textbf{x},\textbf{y}),
\end{equation}
in which $\mathcal{D}=\cup_{n=1}^N\mathcal{D}_n$ is a centralized data set including all users' data sets, $V(\textbf{w}, \textbf{x},\textbf{y})=\frac{1}{N}\sum_{n=1}^N V_n(\textbf{w}, \textbf{x},\textbf{y})$ is a cost function, and $\textbf{W}$ is a closed convex set with a Cartesian product structure, e.g., $\textbf{W} = \prod_{n=1}^N\textbf{W}_n \subseteq \mathbb{R}^{n}$ \citep{7113892}. 
Some examples of the cost functions include 1) Linear regression $V_n(w) = \frac{1}{2}(\textbf{x}_n^T\textbf{w}_n-\textbf{y}_n )^2$, $y_n \in \mathcal{R}_n$; 2) Logistic regression:  $V_n(w) = -\log{(1+\exp{(-\textbf{y}_n\textbf{x}_n^T\textbf{w}_n)})} $, $y_n \in \{-1, 1\}$; 3) Support vector machines: $V_n(w) = \max \{0, 1-\textbf{y}_n\textbf{x}_n^T\textbf{w}_n \}$; and 
4) Complex and non-convex $V_n(w)$ related to neural networks. 


When the computation and storage facilities are not limited and/or privacy of users is not a concern, to solve \eqref{1}, all users can send their data set to a central server. Then, the server calculates the weights and derives the model for any specific application. This approach can be considered as a fully centralized approach, which will be refereed to as centralized solution counterpart of FL. Here, depending on the nature of the problem \eqref{1}, e.g., non-convex problem, diverse algorithms can be applied \citep{7776948}. The benchmark algorithm to solve \eqref{1} is \textit{GD approach} where \eqref{1} is solved iteratively based on the following generic formulation \begin{equation} \label{GD}
\textbf{w}^{t+1}=\textbf{w}^{t}+ \lambda^t \nabla_{\textbf{w}} V\left(\textbf{w}\right), 
\end{equation}
where $\lambda^t$ is a step-size or learning rate parameter \citep{gfed}. Convergence of GD approach can be proved with the following practical assumptions \citep{7776948}: \textit{a)} each $\textbf{W}_n$ is nonempty, closed, and convex;
\textit{b)} $V\left(\textbf{w}\right)$ is smooth (or Sub-gradient Descent for non-smooth functions), \textit{c)} $\bigtriangledown V$ is Lipschitz continuous on $\textbf{W}$ with constant $L_F$. In this context, to improve the performance and convergence, the regularization factors such as norm functor of weights are added to the cost function in \eqref{1}, norm 2 with the definition of $\|\textbf{w}\|_2=(\sqrt{\sum_{n \in \mathcal{N}}w_n^2)}$ \citep{Fedprox}. 

\begin{figure*}
	\begin{center}
		\includegraphics[width=5.1in]{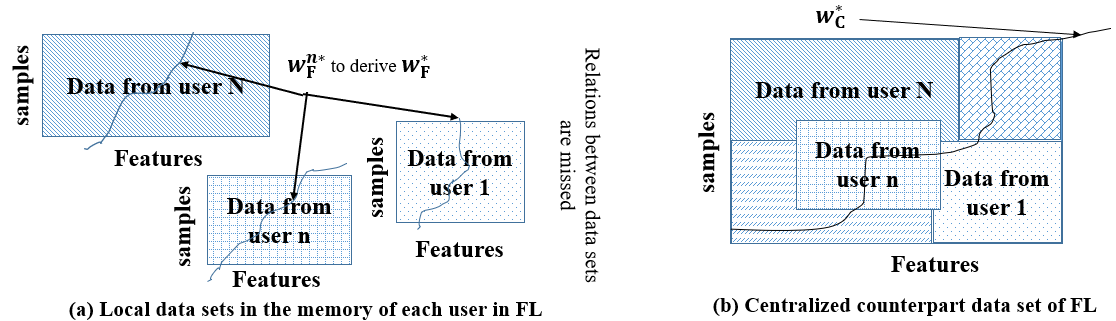}
		\caption{One view of difference of data sets of end-users in FL and centralized counterpart model}
		\label{fig2}
	\end{center}
\end{figure*}

The iterative nature of \eqref{GD} is of high interest for a case where distributed approaches (e.g., FL) are motivated for the following reasons: 1) due to high data volume, solving \eqref{GD} in one node is practically impossible, 2) data holders prefer not to share their own data sets due to privacy and security concerns; 3) data sets are changing fast and sending them to one node needs huge amount of bandwidth; and 4) end users have computational capabilities. All of the above are motivations for \textit{federated learning (FL)}  where \eqref{1} is solved in a distributed manner by end users \citep{gfed}. In FL, each user solves the following optimization problem 
\begin{equation} \label{2}
\min_{\textbf{w}_n \in \textbf{W}_n} V_n(\textbf{w}, \textbf{x}_n, \textbf{y}_n), \,\,\, \forall n \in \mathcal{N}. 
\end{equation}
If data set of each user is $\mathcal{D}_n=\mathcal{D}$\footnote{In the federated learning, the local data sets of users are assumed to be non-i.i.d \citep{gfed}. }, the solution of \eqref{2} ($\textbf{w}^{n*}_{\text{F}}$) can converge to the solution of \eqref{1} without a need to pass information between users. However, due to the limitation to local data set of user, there is a need to message passing algorithms between data end-users in FL \citep{gfed}. In this context, two major architectures are considered for updating the weights between users \citep{FederatedMachineLearning}: 1) \textit{Fusion or server based architecture} where a server sends the model of learning to the users. Each user tries to solve its own problem within a period of time, divided into time slots or epochs. Let assume that the solution of user $n$ in time slot $t$ is $\textbf{w}^{t}_{n}$. Then users send all $\textbf{w}^{t}_{n}$ to the server, and the server updates the weights as follows \citep{gfed}
	\begin{equation}\label{federatedweight}
	\textbf{w}^{t}=\frac{1}{N}\sum_{n=1}^{N}\textbf{w}_n^{t}.
	\end{equation}
Then $\textbf{w}^{t}$ is sent back to all users to utilize for the next step. This algorithm continues until the convergence criteria is met, e.g., $\|\textbf{w}^{t}-\textbf{w}^{t-1}\|_2 \leq \zeta $ where $\zeta$ is a small and positive number. 2)\textit{ Fully decentralized architecture} where there is no server node in the architecture and all users send their weights to each other and calculate the updated weights $\textbf{w}^{t}$ by themselves. 
Each of the two scenarios has its own pros and cons from the implementation perspective, and can be fit to a real scenario \citep{FederatedMachineLearning}. Study of this part is out of the scope of this paper. Our main focus here is how to model the performance loss incurred by the solution of FL e.g., $\textbf{w}^*_{\text{F}}$ and $\textbf{w}^*_{\text{C}}$, which can be defined as 
\begin{equation}\label{distance}
\Delta=\|\textbf{w}^*_{\text{C}}-\textbf{w}^*_{\text{F}}\|_2. 
\end{equation}
We begin by looking at the data sets of users in FL and its centralized counterpart data set which is depicted in Fig. \ref{fig2}. From this figure, it is clear that compared to centralized counterpart, in FL, each user has a truncated version of $\mathcal{D}$, and the impact between data sets of users are missed in the FL algorithm. Consequently, in FL, each user derived solution $\textbf{w}_n^{t}$ from \eqref{2} is uncertain due to its missed knowledge about the relation between its own data set and the other users data sets. Now, the question is how to model this missing information in a such a way that we get a clue about the performance behaviour  of FL. 

\section{Representation of Federated Learning via Worst-case Robust Optimization Theory}

To study the effect of local data sets in FL, we assume that there is an uncertain function
$\textbf{f}_n(\textbf{w}_{-n},\textbf{x}_{-n}, \textbf{y}_{-n})$ that shows the effect of other users data sets on the user $n$ where $\textbf{w}_{-n}$ is a set 
of all weights of all other users except the user $n$, and $(\textbf{x}_{-n}, \textbf{y}_{-n})$ shows all data sets belonging to other users except user $n$. Therefore, FL optimization problem of user $n$ can be represented as  

\begin{equation} \label{2-2}
\min_{\textbf{w}_n \in \textbf{W}_n} \widetilde{V}_n(\textbf{w}, \textbf{x}_n, \textbf{y}_n, \textbf{f}_n(\textbf{w}_{-n},\textbf{x}_{-n}, \textbf{y}_{-n}) ), \,\,\, \forall n \in \mathcal{N}. 
\end{equation}
If each user has a full knowledge of this function, then we again attain the utopia case, i.e., each user can reach to the centralized solution with its own knowledge. Also, if $\textbf{f}_n$ is negligible, $\widetilde{V}_n$ is equal to ${V}_n$. Both of these cases are not practical. The challenge is that it is not straightforward to model this function or even give any distribution function of the uncertain function. However, with the concept of worst case optimization theory, we can assume that there exits a very general information about $\textbf{f}_n(\textbf{w}_{-n},\textbf{x}_{-n}, \textbf{y}_{-n})$ and we can model the actual model of this function as: 
\begin{equation} \label{uncertaintyregion}
 \textbf{f}_n(\textbf{w}_{-n},\textbf{x}_{-n}, \textbf{y}_{-n})= \bar{\textbf{f}}_n(\textbf{w}_{-n},\textbf{x}_{-n}, \textbf{y}_{-n}) +\widehat{\textbf{f}}_n(\textbf{w}_{-n},\textbf{x}_{-n}, \textbf{y}_{-n}), \,\,\, \forall n \in \mathcal{N},
  \end{equation}
where we can consider general form for $\bar{\textbf{f}}_n$, e.g., linear, quadratic or logarithmic function of $\textbf{w}_{-n}$, $\widehat{\textbf{f}}_n$ is the error and uncertain part of this function, and $\textbf{f}_n$ is the actual function. We also assume that the uncertain values for this function is trapped into the bounded and closed region as 
\begin{equation} \label{uncertainityregion}
\Re_n (\textbf{w}_{-n})=\{ \|\widehat{\textbf{f}}_n(\textbf{w}_{-n},\textbf{x}_{-n}, \textbf{y}_{-n})\|_p\leq \varepsilon_n\}, \,\, \forall n \in \mathcal{N},
  \end{equation}
where $\|.\|_p$ is the norm function $p$ and $0<\varepsilon_n$ \citep{protectionfunction1,saeedehbook}. Obviously when the data set increases, the cost function decreases. Therefore, via the concept of worst case robust optimization theory, \eqref{2-2} can be rewritten as 
 \begin{equation} \label{wcfederatedlearning}
\min_{\textbf{w}_n \in \textbf{W}_n} \max_{\widehat{\textbf{f}}_n  \in \Re_n(\textbf{w}_{-n}) } \widetilde{V}_n(\textbf{w}, \textbf{x}_n, \textbf{y}_n, \textbf{f}_n(\textbf{w}_{-n},\textbf{x}_{-n}, \textbf{y}_{-n}) ), \,\,\, \forall n \in \mathcal{N}. 
\end{equation}
Since there is no exact model for function $\textbf{f}_n$ and the uncertainty region, we cannot directly solve \eqref{wcfederatedlearning}. Instead, we will show how we can use the solution of centralized solution via VI to get insight about the solution of \eqref{wcfederatedlearning} and the performance gap, i.e., $\Delta$. 



\subsection{Study of the Centralized Counterpart Solution  }
\eqref{1} which is the centralized solution counterpart of FL can be represented based on VI \citep{6619406,PangVI}. Consider the mapping vector $\mathcal{F}\left(\textbf{w}\right)=\left(\mathcal{F}_{n}\left(\textbf{w}\right)\right)_{n=1}^{N}$, where
\begin{equation}\label{mappingG}
\mathcal{F}_n\left(\textbf{w}\right)= \nabla_{\textbf{w}} V_n\left(\textbf{w}\right),
\end{equation}
in which $\nabla_{\textbf{w}} V_n(\textbf{w})$ denotes the column gradient vector of $ V_n(\textbf{w})$ with respect to
$\textbf{w}$. The solution of \eqref{1} can be obtained by solving $VI(\textbf{w}, \mathcal{F})$ (Proposition 1.4.2 in \citep{PangVI}) as $(\textbf{w}-\textbf{w}^*)\mathcal{F}(\textbf{w}^*) \geq 0$, for all $\textbf{w} \in \textbf{W}$. For \eqref{mappingG}, 1) If $\mathcal{F}$ is monotone on $\textbf{W}$, the solution set of $VI(\textbf{W}, \mathcal{F})$ is closed and convex (possibly empty). 2) If $\mathcal{F}$ is strictly monotone on $\textbf{W}$, $VI(\textbf{W}, \mathcal{F})$ has at most one solution. 3) If $\mathcal{F}$ is strongly-monotone on $\textbf{W}$, $VI(\textbf{W}, \mathcal{F})$  has a unique solution.

When $V_{n}(\textbf{w}_{n},\textbf{f}_n)$ is a continuous 
with respect to $\textbf{w}_{n}\in\textbf{W}_n$, $\mathcal{F}(\textbf{w})$ is a continuous mapping. Therefore, the solution set of $VI(\textbf{W},
\mathcal{F})$ is a closed set  
\citep{PangVI}). To study the federated learning from \eqref{wcfederatedlearning},  define the mapping $\mathcal{F}(\textbf{w})$ where $\alpha_{n}(\textbf{w})\triangleq \text{smallest eigenvalue of}   \nabla^{2}_{\textbf{w}_{n}}V_{n}(\textbf{w})$, $
\beta_{nm}(\textbf{w}) \triangleq \| \nabla_{\textbf{w}_{n}\textbf{w}_{m}} V_{n}(\textbf{w})\|_2$ for all  $n\neq m$,
where $\nabla^{2}_{\textbf{w}_{n}}
V_{n}(\textbf{w})$ and $\nabla_{\textbf{w}_{n}\textbf{w}_{m}}
V_{n}(\textbf{w}))$ are the $K \times K$ Jacobian matrices of
$\mathcal{F}_n(\textbf{w})$ with respect to $\textbf{w}_n$ and
$\textbf{w}_m$, respectively, and $\|- \nabla_{\textbf{w}_{n}\textbf{w}_{m}} V_{n}(
\textbf{w}_n)\|_2$ is the $l_2$-norm of
$V_{n}(\textbf{w}_{n})$. Also, consider
$\label{alphamin}\alpha_{n}^{\text{min}} \triangleq
\inf_{\textbf{w} \in \textbf{W}} \alpha_{n}(\textbf{w})$, $\beta_{nm}^{\text{max}} \triangleq 
\sup_{\textbf{w} \in \textbf{W}}\beta_{nm}(\textbf{w})$,
and 
\begin{eqnarray} \label{Upsilon}
[ \Upsilon]_{nm}= \left\{\begin{array}{c}
\alpha_{n}^{\text{min}}, \,\,\,\,\,\qquad\text{if} \qquad\, m=n, \\
-\beta_{nm}^{\text{max}},  \,\qquad  \text{if} \qquad\, m\neq n. \\
\end{array} \right.
\end{eqnarray}
When $\boldsymbol{\Upsilon}$ is a $P$-matrix (i.e., for any nonzero vector $\textbf{x}$, we have $x_n(\boldsymbol{\Upsilon}\textbf{x})_n>0$ where $x_n$ is the $n^{\text{th}}$ element of $\textbf{x}$), there is the unique solution for \eqref{1}. In the following, we will show how  this condition can be used to study the solution of  \eqref{wcfederatedlearning} and $\Delta$.

\subsection{Study the Federated Solution Via Worst Case Robust Optimization Theory}
Now, we study how FL solution from \eqref{wcfederatedlearning} can be reformulated based on VI. We are looking for the mapping and variable space of FL problem. For \eqref{wcfederatedlearning}, we can define a mapping $\widetilde{\mathcal{F}}\left(\textbf{w}\right)=\left(\widetilde{\mathcal{F}}_{n}\left(\textbf{w}\right)\right)_{n=1}^{N}$ where
\begin{equation}\label{mappingG}
\widetilde{\mathcal{F}}_n\left(\textbf{w}\right)= \nabla_{\textbf{w}} \widetilde{V}_n\left(\textbf{w}\right),
\end{equation}
in which $\nabla_{\textbf{w}} \widetilde{V}_n(\textbf{w})$ denotes the column gradient vector of $\widetilde{V}_n(\textbf{w})$ with respect to
$\textbf{w}$. For FL problem, the domain of the optimization problem is changed for each user $n$, and we have $\widetilde{\textbf{W}}_n(\textbf{w}_{-n},\textbf{a}_{-n},\textbf{b}_{-n} )=\textbf{W}_n \times \Re_{n}(\textbf{f}_{n}) $ which is a function of other users weights and data sets. Therefore, FL problem should be represented by generalized VI (GVI) as $GVI(\widetilde{\textbf{W}}, \widetilde{\mathcal{F}})$ where $\widetilde{\textbf{W}}= \prod_{n \in \mathcal{N}}\textbf{W}_n \times \Re_{n}(\textbf{f}_{n})$  \citep{6619406}. 
To solve \eqref{wcfederatedlearning}, first we need to show that $\widetilde{\textbf{W}}_n$ is a convex, bounded, and closed set and there is a relation between $\widetilde{\mathcal{F}}_n$ and $\mathcal{F}_n$. As a first step, note that we can rewrite the inner optimization problem in \eqref{wcfederatedlearning} as
\begin{equation}\label{psi2}
\psi_n (\textbf{a}_n, \textbf{a}_{-n})= \max_{\widetilde{\textbf{f}}_{n}\in \Re_{n}(\textbf{w}_{-n})} \widetilde{V}_{n}(\textbf{w}_n, {\textbf{f}}_n) =\widetilde{V}_{n}(\textbf{w}_n, \widetilde{\textbf{f}}^{*}_{n}),
\end{equation}
where $\widetilde{\textbf{f}}^{*}_{n}=\textbf{f}_n- \varepsilon_n \boldsymbol{\vartheta}_{n}$ and
$\boldsymbol{\vartheta}_{n} =\frac{\frac{\partial \widetilde{V}_n(\textbf{w}_n, {\textbf{f}}_n)}{\partial \widetilde{\textbf{f}}_n}}{\|\frac{\partial \widetilde{V}_n(\textbf{w}_n, {\textbf{f}}_n)}{\partial {\textbf{f}}_n}\|_2}$. 
$\psi_n (\textbf{a}_n, \textbf{a}_{-n})$ is a concave and continuous differentiable function of $\textbf{a}_n$ for every $\textbf{a}_{-n}$ \citep{6619406}. 
From this reformulation, it can be shown that FL has a solution for any data set, $\widetilde{\textbf{W}}_n$, and $\varepsilon_n$ for all $n \in \mathcal{N}$ (Theorem 1 in \citep{6619406}); and the mapping
$\widetilde{\mathcal{F}}(\textbf{w})$ is a bounded perturbed version of the mapping $\mathcal{F}(\textbf{w})$, i.e., there exists a $0<\wp<\infty$ such that $\|\widetilde{\mathcal{F}}_n(\textbf{w})-
\mathcal{F}_n(\textbf{w})\|_2\leq \wp$. Now, based on this derivatives, we can analyze the solution of FL based on the centralized solution as follows.

\textbf{Remark 1.} When $\boldsymbol{\Upsilon}$
in \eqref{Upsilon} is a $P$-matrix, for any bounded $\boldsymbol{\varepsilon}=[\varepsilon_1,\cdots,\varepsilon_N]$ (Theorem 2 in \citep{6619406}): \textbf{1)} the solution of FL is unique; \textbf{2)} We have $\textbf{w}^*_{\text{C}}\leq\textbf{w}^*_{\text{f}}$; \textbf{3)} The distance between FL and the centralized solutions can be derived  from 
\begin{equation}
\label{upperbound of variations}
\Delta=\|\textbf{w}^*_{\text{C}}-\widetilde{\textbf{w}}^*_{\text{F}}\|_2\leq \frac{\|\boldsymbol{\varepsilon}\|_2}{c_\text{sm}(\mathcal{F})},
\end{equation}
where $c_\text{sm} > 0$ is the strong monotonicity constant for the mapping $\mathcal{F}$, which guarantees $(\textbf{w}_1-\textbf{w}_2)(\mathcal{F}(\textbf{w}_1)-\mathcal{F}(\textbf{w}_2))\geq
c_\text{sm} \|\textbf{w}_1-\textbf{w}_2\|^2_2$ for all
$\textbf{w}_1$,$\textbf{w}_2$ $\in \textbf{W}$ \citep{PangVI}.

Remark 1 shows when the cost function is convex, the federated solution cannot outperform the centralized solution and the gap of the performance can be predicted. Even, based on the concept of local sensitivity analysis of VI, and the definition of $\widetilde{\mathcal{F}}(\textbf{w})$, one can show that for small values of $\varepsilon_n$, this statement is true even for the non-strict monotone maps (Section 5 in \citep{PangVI}). However, for the case that the cost function is non-convex, there is not any solution except for very limited scenarios 
\citep{Cannelli2019}. In this case, when the $\varepsilon_n$ is large and/or the initial values are different, FL approach has the chance to explore the non-convex feasibility set of \eqref{1}, and converges to a solution which can outperform the solution of counterpart centralized solution as reported in \citep{McMahan2016CommunicationEfficientLO}. 

\textbf{Example with Definition of Protection Function: }Here, with one example, we show how with the definition of protection function, the robust problem in \eqref{wcfederatedlearning} can be transformed into a more tractable formulation. Consequently, it can be solved in a more straightforward manner. Consider a linear regression cost function, i.e., $V_n=\frac{1}{2} (\textbf{x}_n^{T}\textbf{w}-\textbf{y})^2$ and assume that the $f_n$ is also linear function of other users weights and data sets. Also, we consider that the uncertainty region for user $n$ is 
$\Re_{n}= \{ \textbf{f}_n= \frac{1}{2} (\textbf{x}_{-n}^{T}\textbf{w}_{-n}-\textbf{y}_{-n}) \, | \, 
\|\textbf{x}_{-n}\|_p\leq \varepsilon_n, + \|\textbf{y}_{-n}\|_{p} \leq \delta_{n} \}$. For this example, based on the concept of protection function \citep{protectionfunction,protectionfunction1}, we can rewrite the cost function  $\widetilde{V}_n(\textbf{w}_n, \textbf{x}_n,\textbf{y}_n, \widetilde{\textbf{f}}_{n}(\textbf{x}_{-n},\textbf{y}_{-n}))$ as


$$\widetilde{V}_n=\frac{1}{2} (\textbf{x}_n^{T}\textbf{w}_n-\textbf{y}_n)+\underbrace{\max_{\widehat{\textbf{f}}_n \in \Re_{n}}\frac{1}{2} (\textbf{x}_{-n}^{T}\textbf{w}_{-n}-\textbf{y}_{-n})}_{\text{protection function}}, \, \, \forall n \in \mathcal{N}, $$
where the last term of the above can be considered as a protection function which can be rewritten as
$\widetilde{V}_n=\frac{1}{2} (\textbf{x}_n^{T}\textbf{w}_n-\textbf{y}_n)+\varepsilon_n\|\textbf{w}_{-n}\|^*_p-\delta_n $ where $\|\textbf{x}|\|^*_p$ is the dual norm of $\|\textbf{x}|\|_p$. Recall that for the linear norm, the dual norm is norm with with order $q=1+\frac{1}{p-1}$. For instance, for norm $2$, we have 
\begin{equation}\label{linearworst}
\widetilde{V}_n=\frac{1}{2} ((\textbf{x}_n^{T}\textbf{w}_n-\textbf{y}_n)^2+ \underbrace{\varepsilon_n\|\textbf{w}_{-n}\|_2+\delta_n}_{\text{protection function}}), \,\, \,\, \forall n \in \mathcal{N}.
\end{equation}
Note that \eqref{linearworst} can be solved with the traditional FedAvg, FedProx and SGD \citep{gfed} where  the protection function is rewritten as $\varepsilon_n\|\textbf{w}-\textbf{w}_n\|_2+\delta_n$\footnote{Even with considering the worst case robust optimization theory, the  server needs to pass the average of weights of all users in each round of iteration}.
Consequently, the robust presentation of \eqref{2} can reach to the tractable solution and does not involve in high computational complexity. Also for the additively coupled structure of cost function and uncertain functions, e.g, logarithmic, exponential, fractional functions, \eqref{2} can be again simplified to more tractable formulations \citep{6619406,saeedehbook}.

This tractable formulation in \eqref{linearworst} also has an interesting interpretation. Via the concept of protection function and worst case approach, the optimization problem of each user in \eqref{linearworst} is augmented with the weights and some positive values, i.e., $\varepsilon_n$ and $\delta_n$, where we will investigate their effects later. This is similar to works in which the perturbed functions or the regularization parameters are added to the objective function of each user in FL and that report the better performance for FL \citep{Fedprox}. Therefore, this representation can help to understand the behavior of proposed algorithms in this context based on the regularization factors.    


\subsection{Proximal based Solution for Federated Learning (Proxi-Fed)} Now, we present the proximal based solution for \eqref{wcfederatedlearning}, which is referred to \textit{Proxi-Fed}. Proximal point method is one of a promising iterative approach to solve the distributed problems that involve set-valued mappings 
(Section 12 in \citep{PangVI} and Section 12.6.1 in \citep{Palomar2010}). 
To adapt this approach for our proposed problem, define
$\textbf{b}=[\textbf{b}_1,\cdots,\textbf{b}_N] \in \mathcal{A}$ and $\widehat{\textbf{w}}(\textbf{b})=[\widehat{\textbf{w}}_1(\textbf{b}),\cdots,\widehat{\textbf{w}}_N(\textbf{b})]$
are the solution to the following optimization problem (Section 12.6.1 in \citep{Palomar2010})
\begin{equation}\label{proximalresponsemap}
\widehat{\textbf{w}}(\textbf{b})=\text{argmin}_{\textbf{w} \in \textbf{W}} \Big[\sum_{n=1}^{N} \widetilde{V}_n(\textbf{w}_n,\textbf{b}_{-n})+\frac{1}{2}
\|\textbf{w}-\textbf{b}\|_2^2\Big],
\end{equation}
where $\widehat{\textbf{w}}(\textbf{b})$ is a proximal response map.
Now, \eqref{proximalresponsemap} can be decomposed into $N$ subproblems (one for each user) \citep{Palomar2010}
\begin{equation}\label{proximalresponsemapn}
\widehat{\textbf{w}}_n(\textbf{b})=\text{argmin}_{\textbf{w}_n \in \mathcal{A}_n} \Big[\widetilde{V}_n(\textbf{w}_n,\textbf{b}_{-n})+\frac{1}{2}
\|\textbf{w}_n-\textbf{b}_n\|_2^2\Big], \,\, \forall n \in \mathcal{N}.
\end{equation}
Assume that $\widehat{\textbf{w}}_n(\textbf{b})$ and $\textbf{b}_n$ be the solutions for user $n$ in its current and previous iterations, then a distributed iterative algorithm for FL based on \eqref{wcfederatedlearning} can be developed which is summarized in Table 1.
The distributed algorithm based on the proximal approach converge when $\boldsymbol{\Upsilon}$ in \eqref{Upsilon} is a $P$-matrix, and $ \frac{\partial^3 v_{n}^{k}(w^k_{n}, f_{n}^{k})}{\partial^2 w^k_n \partial f^k_n}=\frac{\partial^3 v_{n}^{k}(w^k_{n},
	f_{n}^{k})}{\partial w^k_n \partial^2 f^k_n}=0$ \citep{6619406}\footnote{This condition does not add any additional constraints for the case that the $\textbf{f}$ is the linear function.}.

\begin{table}[]
	\caption{Proximal Point Method for Federated Learning (Proxi-Fed)} \centering \vspace{-0.0 in}
	\begin{tabular}{l}
		\hline \hline \textbf{Inputs for Each User}:  $\mathcal{N}=[1,\cdots,N]$: Users' iterations, $\varepsilon_n$, and  $0<\zeta<< 1$
		\\ \textbf{Initialization} For $t=0$: Set an initial $\textbf{w}_n(0) $ and a random $\textbf{w}^0$ for all $n \in \mathcal{N}$,
		\\ \textbf{Iterative Algorithm}: For $t=1,\cdots,T$ and $1\gg T$, for all $n \in \mathcal{N}$, Update\\ $\quad \quad \quad \widehat{\textbf{w}}_n^t = \text{argmin}_{\textbf{w}_n \in \textbf{W}_n} \widetilde{V}_n(\textbf{w}_n,\textbf{f}_{-n}^{t-1})+ \frac{1}{2} \|\textbf{w}_n-\widehat{\textbf{w}}_n^{t-1}\|_2^2$ for all $n \in \mathcal{N}$ , \\
		If $\|\textbf{w}^{t-1}-\textbf{w}^{t}\|_{2}\leq \zeta$, End; Otherwise $t=t+1$, continue;
		\\ \hline \hline
		\vspace{-0.0 in}
	\end{tabular}\label{distributedalgorith}
\end{table}

For the linear regression problem solved in \eqref{linearworst}, the FexProxi cost function is 
\begin{equation}\label{proxilinear}
V_n^{\text{Proxi-Fed}}=\frac{1}{2} (\textbf{x}_n^{T}\textbf{w}_n^{t}-\textbf{y}_n)+\underbrace{\varepsilon_n \|\textbf{w}^{t}-\textbf{w}_n\|_p^*+\delta_n+ \frac{1}{2} \|\textbf{w}_n^{t}- \textbf{w}_n^{t-1} \|_2}_{\text{Regularization factors}}, \,\,\,\, \forall n \in \mathcal{N}. 
\end{equation}
Again \eqref{proxilinear} can be solved by the conventional server based update in this context \citep{gfed} with minor modifications. Note that 
\eqref{proxilinear} is augmented via the weights of each user as well as the other users and $\varepsilon_n$ and $\delta_n$. Also, interestingly, comparing \eqref{proxilinear} with the past regularization methods in FL, e.g., \citep{Fedprox}, highlights that with representation of FL via worst case optimization theory and proximal response method, the utilization of regularization factor in this context can be supported. Besides, with different definition of uncertainty region, we obtain a different regularization factor. Therefore, this representation mathematically supports why we need to add regularization factors in most scenarios of federated approaches to reach the better solutions. 

\section{Evaluation Results}
To evaluate the performance of the representation of federated learning based on worst case robust optimization theory, we examine the performance of our proposed approach using the MNIST data set \citep{cohen2017emnist}. This data set generates non-iid data set for each user which is the main assumptions for the FD local data sets., i.e., $\varepsilon_n\|\textbf{w}-\textbf{w}_n\|_p+\delta_n$. This data set can model the non-i.i.d feature for the data sets in federated learning. Note that in this case we have 784 input neurons and 10 output neurons for each user digit. In our setup, there are $20$ users per each epoch who are attending to the FL algorithms, with $\varepsilon=\delta$, $p=2$ for L2, and $p=1$ for L1.  

\begin{figure*}
	\begin{center}
		\includegraphics[width=3.9 in]{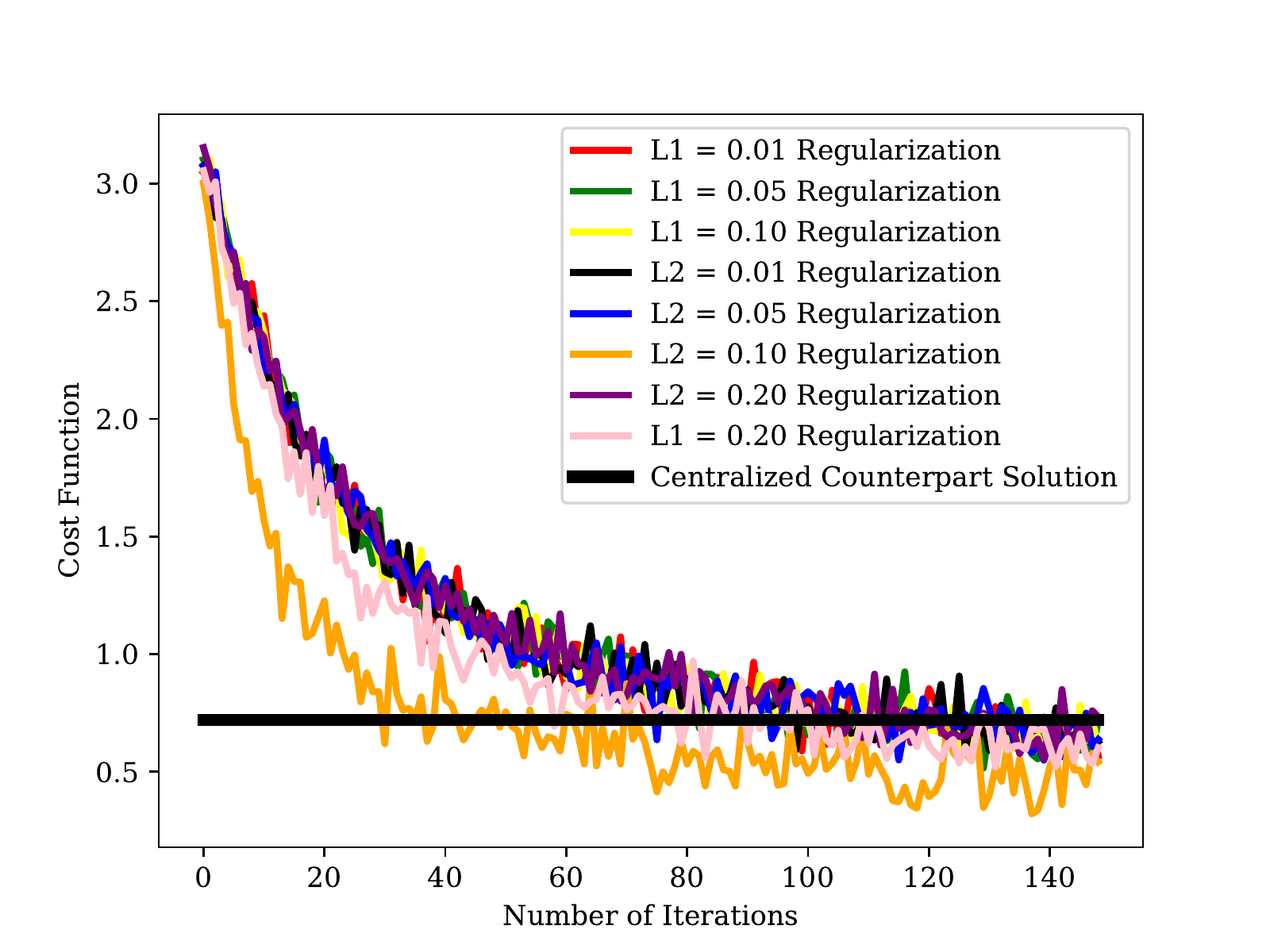}
		\caption{Cost function vs. the number of iterations and the protection function parameters ($p$ and $\varepsilon$)}
		\label{figsimul}
	\end{center}
\end{figure*}

In Fig. \ref{figsimul}, the cost function using norm $2$ (L2) and norm $1$ (L1) versus iterations is depicted for different values of $\varepsilon_n$ and $\delta$ in the training phase. As a base line, we compare the result of FL with its centralized counterpart solution (black curve in Fig. \ref{figsimul}). Since the iteration numbers of the federated learning and its centralized solution are not comparable, we plot the final solution for the centralized counterpart problem. The results in Fig. \ref{figsimul} highlight that $L2$ with $\varepsilon=0.1$ outperforms the other parameter setup for this data set. The interesting point is that for all $L2$ and $L1$ with different $\varepsilon=0.2$, the converged results of federated learning with this protection function is almost better than that of centralized counterpart. In Table \ref{Table}, we show the accuracy of the training set versus $\varepsilon$ for $L2$ and $L1$ norms in protection function in different iteration number. From this results it is clear that $L2$ with $\varepsilon=0.1$ has a fast convergence rate, i.e., its accuracy in $40$ iteration is above 80\%. All other experiments with different protection functions almost approaches 85/\% or more after $120$ iterations. Note that the accuracy of the centralized solution counterpart of our experiment is around $50\%$. This results supports our discussions after Remark 1 that using the federated learning with different regularization factors allow to search the all non-convex feasible set of problem \eqref{1} and there is a chance to converge to a better solution compared to the centralized counterpart problem. Also, there is not any linear relation between increasing the uncertainty bound and decreasing or increasing the cost. This results are valid for this data set and for other data sets, the experiment should be run and the best value of $p$ and $\varepsilon_n$ should be determined. 

\begin{table}
\begin{center}
	\label{Table}
	\caption{Accuracy versus Protection Function Parameters in Federated learning }
\begin{tabular}{cccccccc} \toprule
	{Iterations} & {L1, $\varepsilon=0.01$} & {L1, $\varepsilon=0.1$} & {L1, $\varepsilon=0.2$} & {L2, $\varepsilon=0.01$} & {L2, $\varepsilon=0.1$} & {L2, $\varepsilon=0.2$}\\ \midrule
$40$  & 62\% & 63\% & 65\% & 68\% & 76\% & 60\% 
	    \\ \midrule
	$80$  & 77\% & 74\% & 74\% & 73\% & 77\% & 73\%  \\

	 \midrule
	$120$  & 74\% & 81\% &83\% & 80\% & 89\% & 77\%    \\
 \bottomrule
\end{tabular}
\end{center}
\end{table}
\section{Conclusion}
In this paper, we show how worst-case robust optimization theory can deploy to realize the performance of federated learning compared to its counterpart centralized approaches. Via the concept of protection function, we discuss how different types of regularization parameters can be added to the federated learning problem of each user. We also show how  proximal response map can be deployed in this scenario. By using MNIST data set, we evaluate the proposed reformulation for federated learning which shows the superiority of applying norm $2$ (L2) for this data set. 

\medskip

\small
\bibliographystyle{unsrtnat}

\bibliography{myref}

\end{document}